\pdfoutput=1

\documentclass[11pt]{article}

\usepackage{amsfonts}       
\usepackage{booktabs}
\usepackage{amsmath}
\usepackage{multirow}

\usepackage[final]{acl}

\usepackage{times}
\usepackage{latexsym}

\usepackage[T1]{fontenc}

\usepackage[utf8]{inputenc}

\usepackage{microtype}

\usepackage{inconsolata}

\usepackage{graphicx}

\title{Domain Adaptation of Foundation LLMs for e-Commerce}

\author{
  \textbf{Christian Herold}\thanks{Correspondence author. Email: cherold@ebay.com.} \quad \textbf{Michael Kozielski} \quad \textbf{Tala Bazazo} \quad \textbf{Pavel Petrushkov} \\
  \textbf{Patrycja Cieplicka} \quad \textbf{Dominika Basaj}\thanks{work done while at eBay} \quad \textbf{Yannick Versley} \quad \textbf{Seyyed Hadi Hashemi} \\ 
  \textbf{Shahram Khadivi}\\
  \vspace{0.5cm} \\ eBay Inc.  \\}

\begin{document}
\maketitle
\begin{abstract}
We present the e-Llama models: 8 billion and 70 billion parameter large language models that are adapted towards the e-commerce domain.
These models are meant as foundation models with deep knowledge about e-commerce, that form a base for instruction- and fine-tuning.
The e-Llama models are obtained by continuously pretraining the Llama 3.1 base models on 1 trillion tokens of domain-specific data.

We discuss our approach and motivate our choice of hyperparameters with a series of ablation studies.
To quantify how well the models have been adapted to the e-commerce domain, we define and implement a set of multilingual, e-commerce specific evaluation tasks.

We show that, when carefully choosing the training setup, the Llama 3.1 models can be adapted towards the new domain without sacrificing significant performance on general domain tasks.
We also explore the possibility of merging the adapted model and the base model for a better control of the performance trade-off between domains.
\end{abstract}

\section{Introduction}

Large Language Models (LLMs) have greatly improved the performance on most natural language tasks, and often show surprisingly good zero-shot generalization to new domains \citep{Singhal2023-vs}.
However, training on a specific target domain is often the means of choice to reach the best tradeoff in terms of scalability, domain knowledge, inference costs, and other factors.

While earlier approaches have trained domain-specific models from scratch \citep{beltagy-etal-2019-scibert,alsentzer-etal-2019-publicly}, the big effort of training competitive LLMs have meant that researchers and practitioners
more commonly use continued pretraining (CPT), see e.g. \citet{gururangan-etal-2020-dont,ke2023cpt}, as
compute requirements grew from using 1024 V100 GPUs
over several days for RoBERTa as a larger BERT-like model \citep{liu20roberta} to the 16,000
H100 GPUs used for recent Llama-3
trainings \cite{DBLP:journals/corr/abs-2407-21783}.

For e-commerce applications such as many seen at eBay,
one could use existing pretrained models, such as Llama-3.1 \cite{DBLP:journals/corr/abs-2407-21783} for their use-cases.
However, these models typically lack specific knowledge about the e-commerce domain.

Instead, we continue training the Llama base models on a large amount of e-commerce data. This way we introduce the domain specific knowledge into the model, while at the same time keeping the general capabilities of the model intact.
This technique is known as \lq{}continued pretraining\lq{} and the training setup has to be carefully balanced to prevent the model from degrading too much in performance on general domain tasks.

In Table \ref{tab:related_work}, we compare recent continuous pretraining works in terms of the domain, the size of the models, as well as the amount of training data.
\begin{table*}[t]
\centering
\begin{tabular}{@{}lccc@{}}
\toprule
Study       & Domain & Model Parameter Count & Total num Tokens \\ \midrule
Minerva \cite{DBLP:conf/nips/LewkowyczADDMRS22} & STEM & 8B, 62B, 540B & 26B-38.5B \\
MediTron \cite{DBLP:journals/corr/abs-2311-16079} & Medicine &  7B, 70B & 46.7B \\
Code Llama \cite{DBLP:journals/corr/abs-2308-12950} & Code & 7B, 13B, 34B, 70B & 520B-1,000B \\
Llemma \cite{DBLP:conf/iclr/AzerbayevSPSMJD24} & Math &  7B, 34B & 50B-55B \\
DeepSeekMath \cite{DBLP:journals/corr/abs-2402-03300} & Math & 7B & 500B \\
SaulLM-7B \cite{DBLP:journals/corr/abs-2403-03883} & Law & 7B & 30B \\ 
SaulLM-{54, 141}B \cite{DBLP:journals/corr/abs-2407-19584} & Law &  54B, 141B & 520B \\ 
HEAL \cite{DBLP:conf/naacl/YuanRNRGZCW24} & Medicine & 13B &  14.9B \\ 
Me-LLaMA \cite{DBLP:journals/corr/abs-2402-12749} & Medicine & 13B, 70B &  129B \\ 
ClimateGPT \cite{DBLP:journals/corr/abs-2401-09646} & Climate & 7B, 13B, 70B &  4.2B \\
Nemotron \cite{DBLP:journals/corr/abs-2407-07263} & General & 15B &  1,000B \\ \midrule
e-Llama (ours) & e-commerce & 8B, 70B & 1,000B \\ \bottomrule
\end{tabular}
\caption{Comparing the scale of recent continued pretraining works with our setting. Most existing works are at a significantly smaller scale, either in terms of model size or in terms of tokens used for training.}
\label{tab:related_work}
\end{table*}
As can be seen, our work is at a significantly larger scale than most existing works, either in terms of model size or in terms of tokens used for training, or both.
We share our insights regarding large-scale model adaptation.
In particular, we compare the model adaptation for models of different sizes and discuss the observed differences in behavior.  
We also explore the possibility of model merging to better control the trade-off between general- and domain-specific knowledge.

The rest of the paper is structured as follows.
In Section \ref{sec:data_plus_eval} we discuss our data mixture and explain our methods of model evaluation with focus on e-commerce specific tasks.
In Section \ref{sec:finding_best_setup} we explain our series of experiments to determine the optimal set of hyperparameters.
In Section \ref{sec:e-llama} we show the performance of the final models and discuss the possibility of model merging to better tune for different domains.


\section{Related Work}

The large cost of training LLMs from scratch has meant that continued pretraining
is very attractive for adapting an existing LLM to new languages or domains.
For example, \citet{minixhofer-etal-2022-wechsel} show that it is possible
to reach competitive results for non-English languages by continuing the pretraining of RoBERTa and GPT-2 models.
They start from an English model with tokenizer modification and reach scores on par with monolingual models trained from scratch on a multiple of the data used. In terms of adaptation to different domains, \citet{gururangan-etal-2020-dont} show that continued pretraining on a target domain helps a RoBERTa achieve better performance on tasks in that domain, even taking into account task-specific fine-tuning.

In terms of larger LLMs, \citet{Singhal2023-vs} show that while zero-shot performance of PaLM finetuned on general-domain instruction data is surprisingly good on medical text, continued training on medical instruction data using parameter-efficient finetuning (PEFT) method can further improve these results.
\citet{DBLP:conf/nips/LewkowyczADDMRS22}
show that continued training on a mix of the original data and mathematical
language from ArXiv and math web pages can boost PaLM's performance on mathematical tasks. More recent papers address the problem of catastrophic
forgetting in continued pretraining which can only partially be mitigated
by using the original data in a portion of the continued pretraining mix:
\citet{ke2023cpt} discuss masking updates to the neurons most instrumental
for general-domain performance, and \citet{wu-etal-2024-llama} show that
growing the model by introducing additional layers, followed by only
training these layers can avoid catastrophic forgetting.
In contrast, newer work such as \citet{ke2025demystifyingdomainadaptiveposttrainingfinancial}
is more centered on an optimal data composition, proposing a mix of continued pretraining data
and mixed-in instruction data.

In the e-commerce domain, \citet{DBLP:journals/corr/abs-2402-08831} as well as
\citet{DBLP:conf/aaai/LiMWHJ0X0J24} focus exclusively on instruction tuning,
while our work is the first to consider continued pretraining on domain-relevant data. For other domains, please refer to Table \ref{tab:related_work}.

\section{Setup, Data and Evaluation}
\label{sec:data_plus_eval}


\subsection{Training Framework and Hardware}
For training, we use the Megatron-LM framework from NVIDIA \cite{DBLP:journals/corr/abs-1909-08053, DBLP:conf/sc/NarayananSCLPKV21}.
Training was conducted using 60 nodes, each having 8 NVIDIA H100 80GB GPUs (a total of 480 GPUs).
The GPUs are connected via NVIDIA NVLink (intra-node) and InfiniBand (inter-node).
The hardware is part of the eBay compute platform.

\subsection{Data}
Regarding training data, we mostly follow \citet{DBLP:journals/corr/abs-2406-12023}.
For general domain data, we use a mixture of web-crawled and smaller but more high quality datasets.
We include 10\% non-English general domain data in the data mix.
Regarding the e-commerce domain, we employ several data sources.
On the one hand, we utilize listings and product reviews from the eBay website, as has been done by \citet{DBLP:journals/corr/abs-2406-12023}.
Furthermore, inspired by \citet{lozhkov2024fineweb-edu}, we train an e-commerce classifier and use it to extract e-commerce specific examples from the Fineweb corpus \cite{penedo2024finewebdatasetsdecantingweb}.
We use this data for 20\% of our e-commerce specific data mixture.

\subsection{Evaluation}
\label{subsec:eval}
We perform evaluation both on general and e-commerce specific tasks.
As a first benchmark, we calculate model perplexity on heldout datasets for general and e-commerce data.

\subsubsection*{General Domain}
For evaluating the model capabilities on the general domain for the English language, we utilize the Natural Language Understanding (NLU) benchmark aggregates (in the following called \textbf{NLU En}) also used by \citet{DBLP:journals/corr/abs-2402-00838} and \citet{DBLP:journals/corr/abs-2406-12023} and calculated using the EleutherAI LM Evaluation Harness \cite{eval-harness}.
Furthermore, we utilize the \lq{}Open LLM Leaderboard 2\rq{} \cite{open-llm-leaderboard-v2} (in the following called \textbf{LLM Leaderboard En}) benchmark, which calculates a re-normalized average of the scores for the BBH \cite{suzgun2022challengingbigbenchtaskschainofthought}, GPQA \cite{rein2023gpqagraduatelevelgoogleproofqa}, MUSR \cite{sprague2024musrtestinglimitschainofthought} and MMLU-PRO \cite{wang2024mmluprorobustchallengingmultitask} benchmarks.\footnote{We exclude IFEval and MATH Lvl 5 benchmarks because the former is only useful for instruction-tuned models and the latter gives very low scores for the base models, especially for the 8B model variants.}
When we report model performance in Section \ref{sec:finding_best_setup}, we average \texttt{NLU En} and \texttt{LLM Leaderboard En} scores.
For the evaluation of the non-English, general domain NLU capabilities we use the same task aggregates as \citet{DBLP:journals/corr/abs-2406-12023} (in the following called \textbf{NLU non-En}).
Furthermore, we utilize the \lq{}Open Multilingual LLM Leaderboard\rq{} \cite{lai2023openllmbenchmark} (in the following called \textbf{LLM Leaderboard non-En}).
In this work we focus on German, Spanish, French and Italian.

\subsubsection*{e-Commerce}
Since existing work, like eCeLLM \cite{DBLP:journals/corr/abs-2402-08831} and EcomGPT \cite{DBLP:conf/aaai/LiMWHJ0X0J24} focuses on evaluation of instruction tuned models, we define a total of 5 novel e-commerce benchmarks for evaluation of foundation models.
All tasks are strongly connected to relevant downstream tasks that we encounter in the e-commerce setting.
They revolve around the listings on an e-commerce website, of which we consider title, category, price and a list of aspect key-value pairs\footnote{An example for an aspect key could be \lq{}Brand\rq{} and a possible aspect-value in this case could be \lq{}Nike\rq{}.}.
Below we list the tasks in detail:
\begin{enumerate}
    \item \textbf{Aspect Prediction (AP)}: Given the title and category of a listing, as well as a specific aspect key, predict the corresponding aspect value.
    \item \textbf{Aspect Prediction Multiple Choice (AP$^{\text{MC}}$)}: Given 4 listings, of which 3 are corrupted by changing at least 1 aspect value, the model has to identify the correct listing.
    \item \textbf{Price Prediction Multiple Choice (PP$^{\text{MC}}$)}: Given 4 listings, of which 3 are corrupted by changing the price at which the item was sold, the model has to identify the listing with the correct selling price.
    \item \textbf{Most Common Aspects (MCA)}: Given a category and an aspect key, the model has to predict the most common aspect values for that key.
    \item \textbf{Most Common Aspects Multiple Choice (MCA$^{\text{MC}}$)}: Given a category and an aspect key, the model is presented with 4 choices for the most common aspect value for that key and has to select the correct one.
\end{enumerate}
We evaluate these tasks for English, German, Spanish, French and Italian.
For all tasks, the final evaluation metric is accuracy.
We give an example for each of the task in Appendix \ref{subsec:eval_examples}.

In order to obtain a strong baseline, we perform a set of experiments where we optimize the number of few-shot examples for the base Llama-3 model, see Appendix \ref{sec:few_shot_exp} for the details.

\section{Finding the best Setup}
\label{sec:finding_best_setup}
In this section we discuss several series of experiments we performed to determine the best setup for continuously pretraining.
For these studies we focus on the English language benchmarks, since we assume the non-English languages will follow the same trend.
Since the 3.1 version of Llama was not released at the time, some experiments utilize Llama-3.0 models instead. 
The final models described in Section \ref{sec:e-llama} are based on Llama-3.1.

\subsection{Learning Rate}
Maybe the most important hyperparameter to consider is the maximum learning rate $LR_{max}$ of the continued pretraining.
Meta have used a $LR_{max}$ of 3.0e-4 and 1.5e-4 for their training of Llama-3.1 8B and 70B respectively \cite{DBLP:journals/corr/abs-2407-21783}.
However, using the same maximum learning rate for continued pretraining might not yield the best results as the model might forget too much information from the previous training or, on the contrary, the model might not learn enough from the new data mixture \cite{DBLP:journals/corr/abs-2308-04014, DBLP:journals/tmlr/IbrahimTGRABLR24}.

There are mainly 2 paradigms in existing work: (i) use the same $LR_{max}$ as for the original pretraining \cite{DBLP:journals/corr/abs-2308-12950, DBLP:journals/corr/abs-2311-16079, DBLP:journals/corr/abs-2402-03300}, or (ii) use a smaller value, typically around 10\% of the original $LR_{max}$ \cite{DBLP:conf/iclr/AzerbayevSPSMJD24, DBLP:conf/nips/LewkowyczADDMRS22, DBLP:journals/corr/abs-2407-19584, DBLP:conf/naacl/YuanRNRGZCW24, DBLP:journals/corr/abs-2401-09646, DBLP:journals/corr/abs-2402-12749, DBLP:journals/corr/abs-2407-07263}.

\begin{table}[]
\centering
\begin{tabular}{@{}lcccc@{}}
\toprule
\multirow{2}{*}{$LR_{max}$} & \multicolumn{2}{c}{ppl ($\downarrow$)} & \multicolumn{2}{c}{benchmark ($\uparrow$)} \\ \cmidrule(l){2-5} 
 & e-com. & general & e-com. & general \\ \midrule
Llama-3.0 & 7.28 & 8.38 & 45.9 & 44.1 \\ \midrule
3.0e-5 & 2.03 & 6.43 & 59.9 & 42.0 \\
3.0e-4 & 2.01 & 6.48 & 58.6 & 40.5 \\
3.0e-3 & 2.15 & 7.70 & 50.6 & 34.5 \\ \bottomrule
\end{tabular}
\caption{Effect of the maximum learning rate of the continued pretraining (1 trillion tokens) of Llama-3.0 8B on the final model performance. Llama-3.0 used $LR_{max}$=3.0e-4.}
\label{tab:lr_study}
\end{table}
We perform a set of experiments to determine the best maximum learning rate.
Since the impact of the learning rate might significantly depend on the amount of data used in training, we decide to compare training runs utilizing the full 1 trillion tokens of data (50\% e-commerce ratio). 
In all cases, the learning rate decays over the course of the full training with a cosine scheduling to the minimum learning rate of 3.0e-6.
We compare the final model performance in terms of perplexity on the heldout test sets, as well as general (average of \texttt{NLU En} and \texttt{LLM Leaderboard}) and e-commerce benchmarks.
The results can be found in Table \ref{tab:lr_study}.

In terms of perplexity, we find that a higher learning rate leads to a slightly better score on the new domain.
However, these improvements do not translate to a better score on the e-commerce specific benchmarks.
At the same time, a higher learning rate leads to more degradation on the general domain benchmarks.
This might be an indication that our general domain data mix is maybe a bit lower quality than what has been used by Meta in the Llama-3 pretraining.
In the end, we decide to use an $LR_{max}$ that is 10\% of the maximum learning rate used in pretraining, i.e. \textbf{3.0e-5 for the 8B model and 1.5e-5 for the 70B model}.

\subsection{Data Weighting}

\begin{table}[]
\centering
\begin{tabular}{@{}lcccc@{}}
\toprule
\multirow{2}{*}{\% e-com} & \multicolumn{2}{c}{ppl ($\downarrow$)} & \multicolumn{2}{c}{benchmark ($\uparrow$)} \\ \cmidrule(l){2-5} 
 & e-com. & general & e-com. & general \\ \midrule
Llama-3.0 & 7.28 & 8.38 & 45.9 & 44.1 \\ \midrule
10 & 2.75 & 6.87 & 55.6 & 43.2 \\
25 & 2.59 & 6.92 & 56.7 & 43.1 \\
50 & 2.47 & 7.00 & 57.5 & 43.3 \\
75 & 2.40 & 7.15 & 57.6 & 43.2 \\ \bottomrule
\end{tabular}
\caption{Effect of the amount of e-commerce data in the continued pretraining (30 billion tokens) of Llama-3.0 8B on the final model performance.}
\label{tab:weight_study}
\end{table}

\begin{figure}[t!]
\centering
  \includegraphics[width=0.98\linewidth]{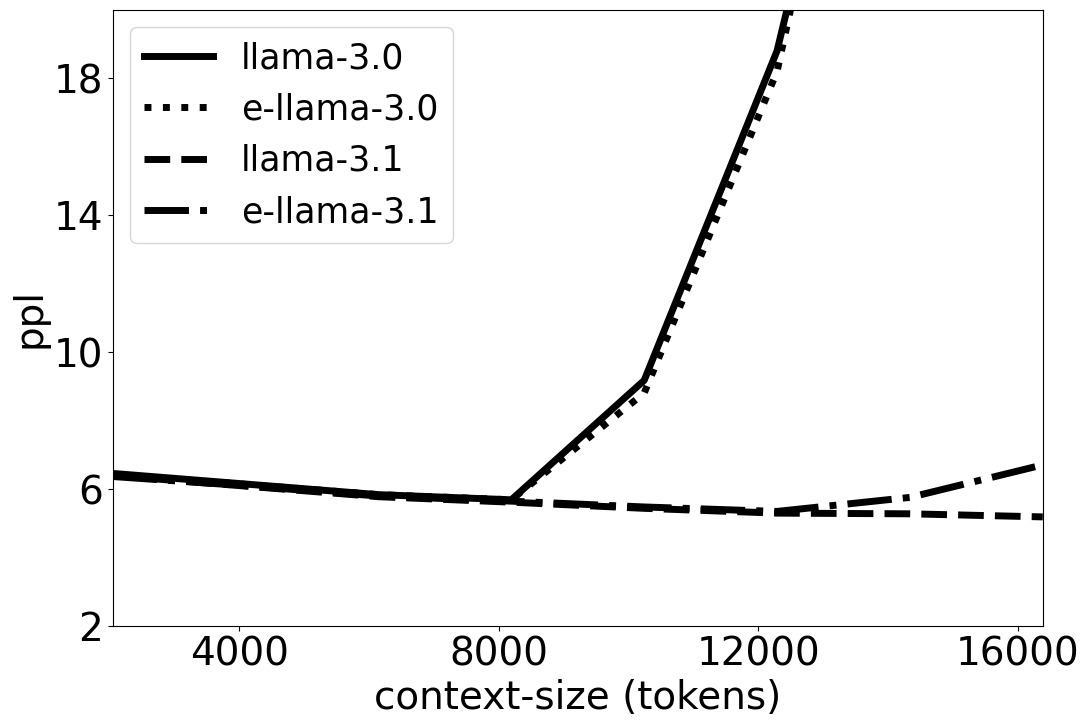}
  \caption {Perplexity as a function of the input sequence length for the 8B (e)-Llama-3.0/3.1 models. The 3.0 variants can not handle context sizes much longer than 8k, since they have never seen these lengths in training.}
  \label{fig:context_size}
\end{figure}

\begin{figure*}[t!]
\centering
  \includegraphics[width=0.295\linewidth]{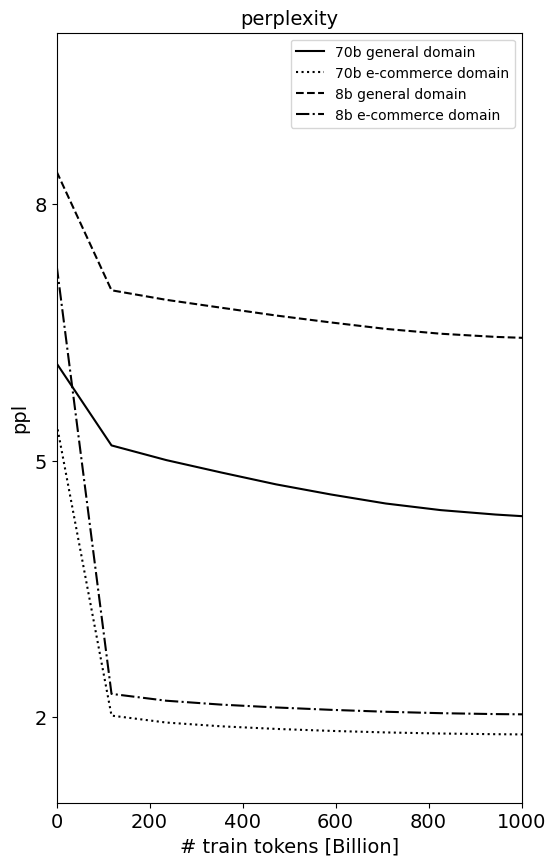} \hfill
  \includegraphics[width=0.32\linewidth]{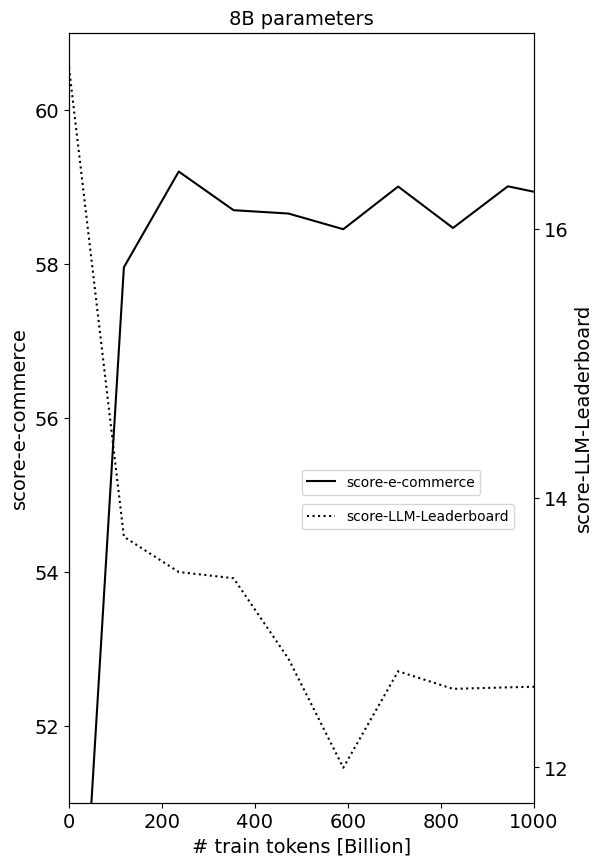} \hfill
  \includegraphics[width=0.32\linewidth]{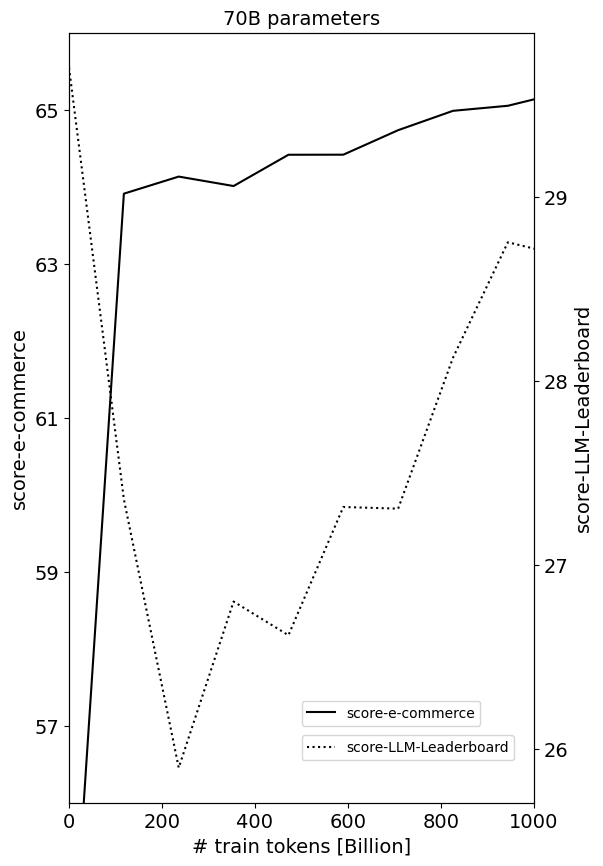}
  \caption {Evolution of the model performance over the course of the training. Left: perplexity on heldout data sets in the general/e-commerce domain. Center/Right: Evolution of downstream model performance on English tasks for 8B/70B model.}
  \label{fig:train_evolution}
\end{figure*}

While we want the model to learn about the new domain, at the same time we want to avoid the effect of catastrophic forgetting.
To combat this, it is common to include some percentage of general domain examples in the data mixture (sometimes called \lq{}replay examples\rq{}).
Most existing works use only up to 15\% of general data in their mixture \cite{DBLP:conf/iclr/AzerbayevSPSMJD24, DBLP:conf/nips/LewkowyczADDMRS22, DBLP:journals/corr/abs-2308-12950, DBLP:journals/corr/abs-2311-16079, DBLP:journals/corr/abs-2403-03883, DBLP:journals/corr/abs-2407-19584} with the exception of \citet{DBLP:conf/naacl/YuanRNRGZCW24} who use 35\%.
However, we have reason to believe that in our specific case, a higher ratio of general domain data might be advisable, because our in-domain e-commerce data comes from a very different distribution.

Following \citet{DBLP:journals/tmlr/IbrahimTGRABLR24} we perform experiments on a limited amount of training tokens (ca 30 billion) with a varying ratio of e-commerce data ($LR_{max}$=3.0e-5).
The results can be found in Table \ref{tab:weight_study}.

As expected, we see that with a higher percentage of e-commerce data, the perplexity and downstream performance for the e-commerce related tasks is improving, although with diminishing returns when going above 50\%.
At the same time, perplexity on the general domain data is getting worse, but this does not effect the model scores on the general domain benchmarks.
As mentioned before, the reason for this is most likely the different distributions of the general domain data we are using vs the one that was used in pretraining.
In the end, we decide to continue pretraining with an \textbf{e-commerce percentage of 50\% in our data mix}.

\subsection{Context Size}

Finally, we explore the effect of the continued pretraining on the context size of the model.
While Llama-3.0 has a context size of 8k, the Llama-3.1 models have a much larger context size of 128k.
Ideally we would like to continue the pretraining with the same large context size, but this introduces several challenges.
First, the vast majority of our training examples both for general and e-commerce domain are shorter than 1k tokens.
Additionally, increasing the context size makes it harder to train the model efficiently due to the quadratic computational complexity of the transformer model.
There exist methods to mitigate the latter issue, like the context parallel training approach \cite{DBLP:journals/corr/abs-2405-07719} but when applying said approach, we find that this still introduces too much computational overhead and significantly slows down the training.
We decide to continuously train both Llama-3.0 8B and 3.1 8B with 8k context size and study the effects this has on the models.
In Figure \ref{fig:context_size} we calculate the perplexity of the 8B base and e-Llama models as a function of the input sequence length for a general domain heldout test set.

All models exhibit nearly identical performance for inputs smaller than 8k.
Unsurprisingly, the 3.0 variants can not handle inputs larger than 8k at all.
The e-Llama model that is based on Llama-3.1 exhibits a much better understanding of longer sequences, even though it has not seen any sequences longer than 8k in the continued pretraining.
We can conclude that the model retains most of its ability to handle longer sequences.
We do see some degradation for even longer input lengths, but this is an acceptable trade-off for us.
We therefore decide to \textbf{perform the continued pretraining with a context size of 8k}.

\begin{table*}[t!]
\centering
\begin{tabular}{@{}lcccccccccc@{}}
\toprule
\multicolumn{1}{c}{\multirow{3}{*}{Model}} & \multicolumn{4}{c|}{general domain benchmarks ($\uparrow$)} & \multicolumn{6}{c}{e-commerce benchmarks ($\uparrow$)} \\ \cmidrule(l){2-11} 
\multicolumn{1}{c}{} & \multicolumn{2}{c|}{En} & \multicolumn{2}{c|}{non-En} & \multicolumn{5}{c|}{En} & \multicolumn{1}{c}{non-En} \\
\multicolumn{1}{c}{} & \multicolumn{1}{c}{NLU} & \multicolumn{1}{c|}{Lead.} & \multicolumn{1}{c}{NLU} & \multicolumn{1}{c|}{Lead.} & \multicolumn{1}{c}{AP} & \multicolumn{1}{c}{AP$^\text{MC}$} & \multicolumn{1}{c}{PP$^\text{MC}$} & \multicolumn{1}{c}{MCA} & \multicolumn{1}{c|}{MCA$^\text{MC}$} & \multicolumn{1}{c}{avg.} \\ \midrule
\textbf{8B} &  & \multicolumn{1}{c}{} &  &  &  &  & &  & \multicolumn{1}{c}{} &  \\
Llama-3.1 & 71.8 & \multicolumn{1}{c}{17.2} & 54.1 & 43.2 & 36.5 & 61.8 & 50.1 & 27.4 & \multicolumn{1}{c}{55.3} & 35.8 \\
e-Llama & 71.6 & \multicolumn{1}{c}{12.6} & 54.0 & 42.4 & 54.9 & 74.9 & 59.6 & 37.8 & \multicolumn{1}{c}{67.4} & 46.8 \\ \midrule
\textbf{70B} &  & \multicolumn{1}{c}{} &  &  &  & &  &  & \multicolumn{1}{c}{} &  \\
Llama-3.1 & 76.6 & \multicolumn{1}{c}{29.7} & 58.5 & 55.2 & 42.8 & 66.3 & 59.3 & 35.2 & \multicolumn{1}{c}{61.9} & 40.4 \\
e-Llama & 76.3 & \multicolumn{1}{c}{28.7} & 59.2 & 55.4 & 59.2 & 79.5 & 65.7 &  49.9 & \multicolumn{1}{c}{71.5} & 52.8 \\ \bottomrule
\end{tabular}
\caption{Final performance of the e-Llama 8B/70B models on general domain and e-commerce specific evaluation benchmarks.}
\label{tab:final_performance}
\end{table*}

\section{e-Llama}
\label{sec:e-llama}

In this section we discuss the training and performance of the final e-Llama 8B and 70B models.

\subsection{Training}

Our setup mostly follows \citet{DBLP:journals/corr/abs-2406-12023} while taking into account our findings from Section \ref{sec:finding_best_setup}.
In particular we use cosine Learning Rate (LR) scheduling with warmup, a batch-size of ca. 11.8 million tokens and 85k total update steps.

In Figure \ref{fig:train_evolution}, we show the evolution of the 8B/70B model performance over the course of the training.
We see that the perplexity on the general domain data is decreasing for both 8B and 70B model.
At the same time, the gap between 8B and 70B stays constant throughout the training.
This indicates that while the distribution of our general domain data is different from the original one, the complexity of the data might be similar.
Perplexity on the e-commerce data is also decreasing but at a much faster rate. 
In the end, the difference in terms of e-commerce perplexity for 8B and 70B model is much smaller than for the base models.

In terms of downstream performance, we find that the 8B model seems to quickly become saturated and performance is no longer increasing after 20\% of the training.
The 70B model on the other hand recovers much better on the general domain tasks, while also continuously improving in the e-commerce domain.
We think this might be due to the much larger model size, that allows the model to better incorporate new information without catastrophic forgetting.
Also, the smaller learning rate for the 70B model might have played a role here.

\subsection{Final models}

In Table \ref{tab:final_performance} we show the final model performance of e-Llama 8B/70B in comparison to the Llama-3.1 base models.

On the general domain \texttt{NLU} benchmarks, the e-Llama models perform the same as the base Llama-3.1 models.
On the more challenging \texttt{LLM Leaderboard} tasks, we see some performance degradation, especially for the smaller 8B model variant.
We think this might be due to a combination of smaller model size and a different data distribution of our general domain data compared to what has been used at Meta.

On the e-commerce benchmarks, the e-Llama models improve relative to the Llama-3.1 base models by around 25\% on English and by around 30\% on non-English benchmarks on average.\footnote{The individual language scores for non-English can be found in Table \ref{tab:ecom_per_lang} in the Appendix.}
Interestingly, the gap between 8B and 70B variant for Llama base model and for e-Llama is roughly the same for the e-commerce tasks, even though in terms of perplexity the e-Llama models are closer together (compare left side of Figure \ref{fig:train_evolution}).
This once again highlights that perplexity and downstream performance do not always follow the same trend.

\subsection{Model Merging}

\begin{figure}[t]
\centering
  \includegraphics[width=0.98\linewidth]{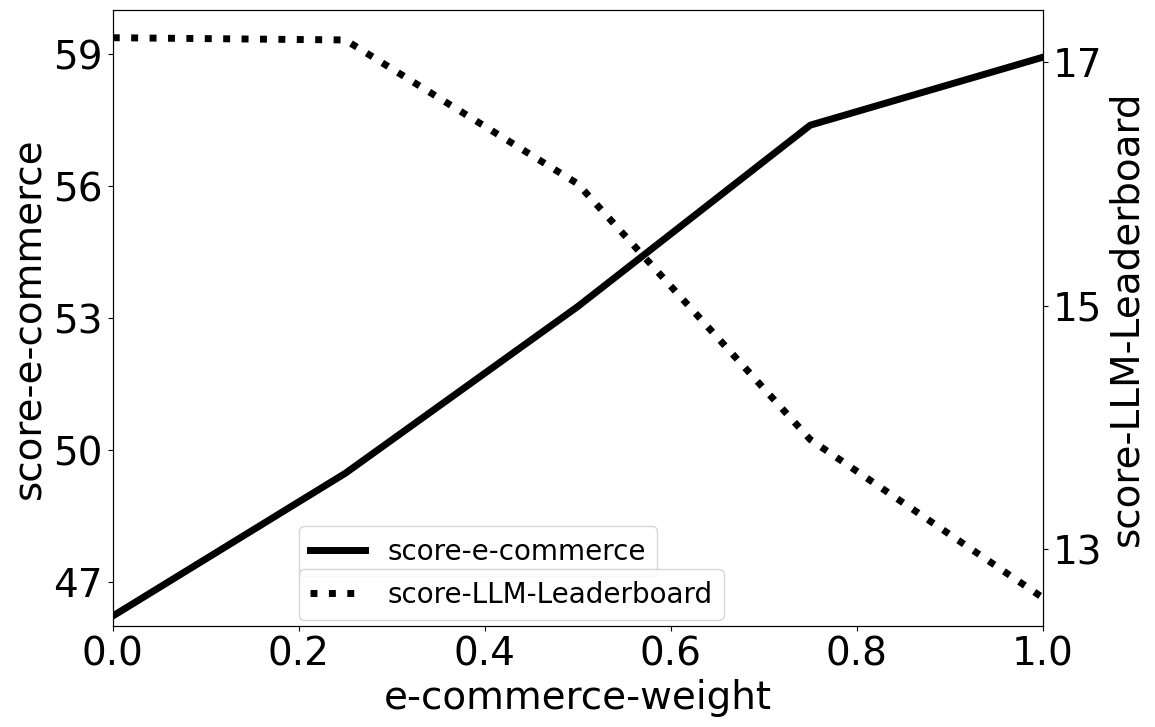}
  \caption {Model merging: 8B Model performance on English general and e-commerce benchmarks as a function of the weight of the e-Llama model parameters vs the base Llama-3.1 model parameters.}
  \label{fig:model_merging}
\end{figure}

The last topic we want to discuss is how to better align the trade-off between general and domain-specific performance.
Lets assume we want to have the best performing model on the e-commerce domain, but we can not allow the general domain performance to drop below a certain threshold on the general domain tasks.
As can be seen from the experiments in Figure \ref{fig:train_evolution} and Table \ref{tab:weight_study}, reducing the percentage or amount of e-commerce training data does not allow to make very precise forecasts of final model performance.
Instead, we utilize a technique called \lq{}model merging\rq{} \cite{DBLP:conf/icml/WortsmanIGRLMNF22}, where we simply average all parameters of the base Llama-3.1 model checkpoint and our final e-Llama model checkpoint.
In Figure \ref{fig:model_merging} we show how the performance of the resulting model changes as a function of the individual model weights.

The performance for both general and e-commerce domain follow an almost linear trend.
This allows for a very precise tuning of the final model performance and has the additional advantage that the model merging is not compute intensive at all.

\section{Conclusion}

We have discussed our efforts to adapt the Llama-3.1 8B and 70B parameter base models towards the e-commerce domain.
In order to evaluate the model capabilities in the e-commerce setting, we design and implement a set of multilingual, e-commerce specific evaluation benchmarks.
Through a series of experiments, we determine the best experimental setting for our use-case.
We show that the models can be adapted well towards the new domain with limited degradation on general domain performance.
Furthermore, we highlight that with model merging, we can very precisely tune the final model performance.

\section{Limitations}

The present work has several limitations:
(i) We focus on a single domain only, namely e-commerce.
(ii) We focus on non-instruction-tuned foundation models only. 
A logical improvement is to consider instruction tuning as an additional part in the pipeline.
(iii) While we try to define a comprehensive set of evaluations for the e-commerce domain, the diversity and quantity of evaluations could be further improved.
(iv) Finally, in this work we focus solely on the Llama family of models.
Future work should explore further open source models.


\bibliography{acl_latex}  

\newpage

\appendix

\section{Appendix}
\label{sec:appendix}

\subsection{Examples for e-commerce tasks}
\label{subsec:eval_examples}
Here, we give an example for each of the e-commerce tasks described in Section \ref{subsec:eval}. All examples are for the English language. For languages other than English, the prompt stays the same, but the item-specific attributes like title are in the corresponding language.
\subsubsection*{AP}
The model has to predict the most probable continuation of the following text input:

\texttt{For an e-commerce website, under the category "Video Games \& Consoles:Video Games", the listing with the title "Dark Souls III (Sony PlayStation 4)" has the following aspect key-value pairs:\\
Rating:}

\subsubsection*{AP$^{\text{MC}}$}
The model is used to independently score the following 4 text sequences, has to give the highest probability to the correct sequence (first one).

\texttt{For an e-commerce website, the listing with the title "Dark Souls III (Sony PlayStation 4)" has the following aspect key-value pairs associated with it:\\
Rating: M - Mature}\\
 
\texttt{For an e-commerce website, the listing with the title "Dark Souls III (Sony PlayStation 4)" has the following aspect key-value pairs associated with it:\\
Rating: E - Everyone}\\
 
\texttt{For an e-commerce website, the listing with the title "Dark Souls III (Sony PlayStation 4)" has the following aspect key-value pairs associated with it:\\
Rating: T - Teen}\\
 
\texttt{For an e-commerce website, the listing with the title "Dark Souls III (Sony PlayStation 4)" has the following aspect key-value pairs associated with it:\\
Rating: AO - Adults Only}

\subsubsection*{PP$^{\text{MC}}$}
The model is used to independently score the following 4 text sequences, has to give the highest probability to the correct sequence (first one).

\texttt{For the listing with the title "Authentic Louis Vuitton Monogram Empreinte Bastille PM 2Way Tote Bag Black 9281E", the final selling price was \$816.00.}\\
 
\texttt{For the listing with the title "Authentic Louis Vuitton Monogram Empreinte Bastille PM 2Way Tote Bag Black 9281E", the final selling price was \$81.60.}\\
 
\texttt{For the listing with the title "Authentic Louis Vuitton Monogram Empreinte Bastille PM 2Way Tote Bag Black 9281E", the final selling price was \$204.00.}\\
 
\texttt{For the listing with the title "Authentic Louis Vuitton Monogram Empreinte Bastille PM 2Way Tote Bag Black 9281E", the final selling price was \$1632.00.}\\

\subsubsection*{MCA}
The model has to predict the most probable continuation of the following text input:

\texttt{For an e-commerce website, under the category "Clothing, Shoes \& Accessories:Women:Women's Clothing:Coats, Jackets \& Vests", the following are the most common aspect values for the aspect key "Outer Shell Material":}

\subsubsection*{MCA$^{\text{MC}}$}
The model is used to independently score the following 4 text sequences, has to give the highest probability to the correct sequence (first one).

\texttt{For an e-commerce website, under the category "Cameras \& Photo:Digital Cameras", the most common aspect value for the aspect key "Brand" is "Canon".}\\
 
\texttt{For an e-commerce website, under the category "Cameras \& Photo:Digital Cameras", the most common aspect value for the aspect key "Brand" is "Fujifilm".}\\
 
\texttt{For an e-commerce website, under the category "Cameras \& Photo:Digital Cameras", the most common aspect value for the aspect key "Brand" is "PENTAX".}\\
 
\texttt{For an e-commerce website, under the category "Cameras \& Photo:Digital Cameras", the most common aspect value for the aspect key "Brand" is "Nikon".}\\

\subsection{Optimizing the Few-Shot Setup}
\label{sec:few_shot_exp}

\begin{figure}[t]
\centering
  \includegraphics[width=0.98\linewidth]{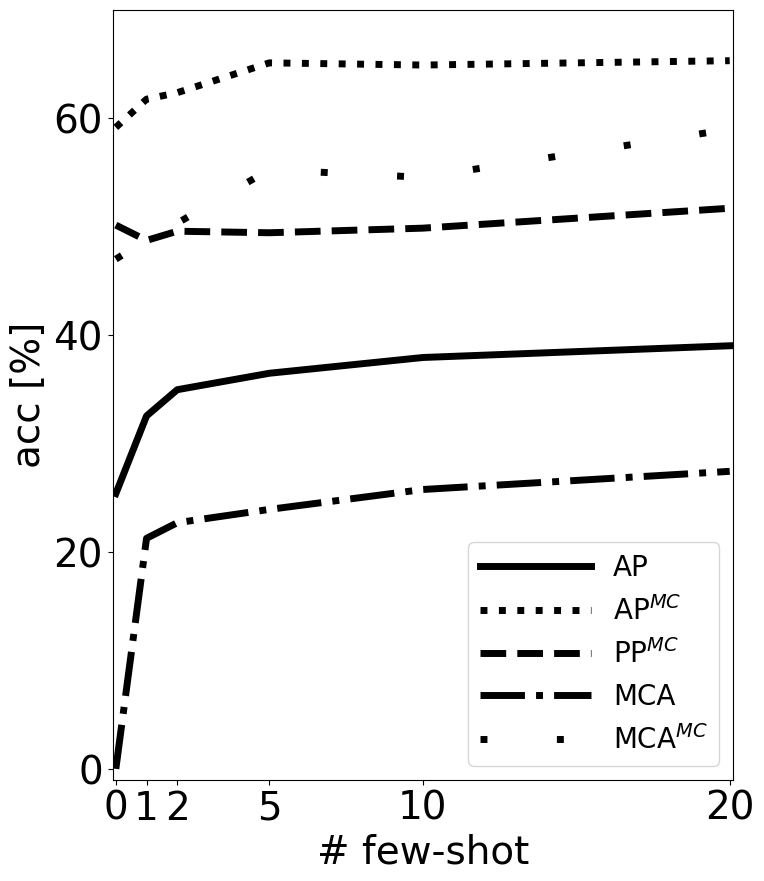}
  \caption {Llama-3 8B model performance on the 5 e-commerce evaluation tasks as a function of the number of few-shot examples provided in the prompt.}
  \label{fig:eval_few_shot}
\end{figure}

For the base Llama-3 8B model, we prompt the model for each of the above tasks with up to 20 few-shot examples.
The results can be seen in Figure \ref{fig:eval_few_shot}.

We find that for the \texttt{PP}$^{\texttt{MC}}$ task, few-shot prompting does not significantly improve the model performance, therefore in the following we use \textbf{0-shot} evaluation for this task.
Since we have already a quite high score for the \texttt{AP}$^{\texttt{MC}}$ task, and there is only limited information to be gained from the few-shot examples, we decide to use \textbf{1-shot} evaluation for this task.
For both \texttt{AP} and \texttt{MCA}$^{\texttt{MC}}$ we see improvements with more few-shot examples. Therefore we end up using \textbf{5-shot} evaluation for these tasks.
Finally for \texttt{MCA} we have quite low scores overall, and the model seems to benefit from more few-shot examples. Therefore we end up using \textbf{20-shot} evaluation for this task.

\subsection{Non-English Benchmark Scores}

\begin{table}[h!]
\centering
\begin{tabular}{@{}lcccc@{}}
\toprule
\multicolumn{1}{c}{\multirow{2}{*}{Model}} & \multicolumn{4}{c}{e-commerce benchmarks ($\uparrow$)} \\ \cmidrule(l){2-5} 
\multicolumn{1}{c}{}                       & De       & Fr       & It       & Es       \\ \midrule
\textbf{8B}                                         &          &          &          &          \\
Llama-3.1                                  & 35.4     & 35.0     & 35.0     & 37.6     \\
e-Llama                                    & 47.5     & 47.0     & 46.0     & 46.7     \\ \midrule
\textbf{70B}                                        &          &          &          &          \\
Llama-3.1                                  & 39.2     & 40.6     & 40.7     & 41.2     \\
e-Llama                                    & 52.7     & 52.2     & 54.5     & 52.0     \\ \bottomrule
\end{tabular}
\caption{Final performance of the e-Llama 8B/70B models on language-specific, e-commerce evaluation benchmarks.\label{tab:ecom_per_lang}}
\end{table}

\end{document}